%% file: main.tex
\documentclass[10pt,twocolumn,letterpaper]{article}

\usepackage{cvpr}              
\usepackage{amsmath}
\usepackage{bbold}
\usepackage{color}
\usepackage{colortbl}

\definecolor{cvprblue}{rgb}{0.21,0.49,0.74}
\usepackage[pagebackref,breaklinks,colorlinks,allcolors=cvprblue]{hyperref}

\def\paperID{13057} 
\def\confName{CVPR}
\def\confYear{2025}

\title{ConMo: Controllable Motion Disentanglement and Recomposition \\ for Zero-Shot Motion Transfer}

\author{
Jiayi Gao$^{1,2 \dagger }$ \quad Zijin Yin$^{2}$ \quad Changcheng Hua$^{1}$ \quad Yuxin Peng$^{1}$ \quad Kongming Liang$^{2}$ \\ \quad Zhanyu Ma$^{2}$ \quad Jun Guo$^{2}$ \quad Yang Liu$^{1 \ast}$ \\
${}^{1}$ Wangxuan Institute of Computer Technology, Peking University\\
${}^{2}$ Beijing University of Posts and Telecommunications \\
{\tt\small hcc@stu.pku.edu.cn, \{pengyuxin, yangliu\}@pku.edu.cn,}\\
{\tt\small \{gaojiayi, yinzijin2017, liangkongming, mazhanyu, guojun\}@bupt.edu.cn}
}

\begin{document}
\twocolumn[{
\maketitle
\renewcommand\twocolumn[1][]{#1}
\vspace{-3em}
\begin{center}
    \centering
    \captionsetup{type=figure}
    \includegraphics[width=\textwidth]{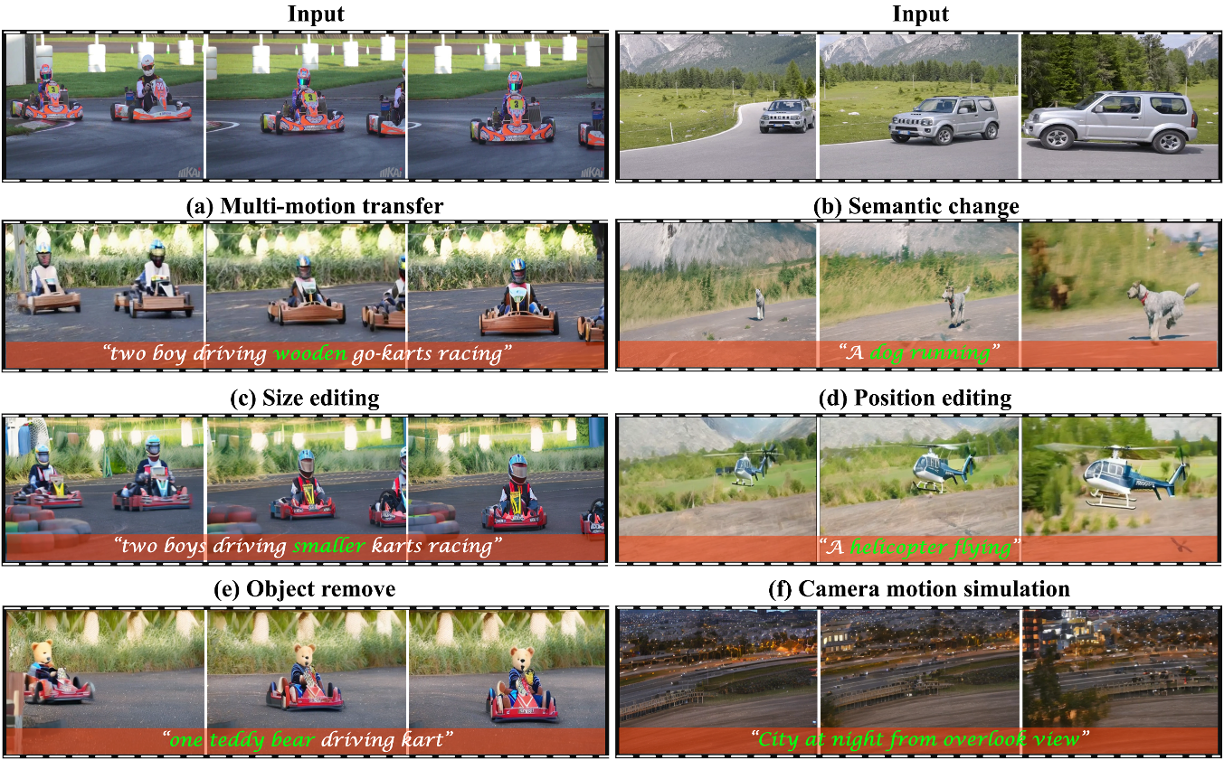}
    \vspace{-1.5em}
    \captionof{figure}{We propose ConMo to achieve various motion transfer applications: (a) multi-subjects motion transfer, (b) subject semantic/category change, (c)  subject size editing, (d) subject position editing, (e) object remove and (f) camera motion simulation.(Green text indicates major changes.)}
    \label{fig:teaser}
\end{center}%
}]

\let\thefootnote\relax\footnotetext{$\dagger$ Jiayi Gao is jointly affiliated with Peking University and Beijing University of Posts and Telecommunications, both recognized as co-primary institutions.}
\let\thefootnote\relax\footnotetext{$\ast$ Corresponding author.}

\input{sec/0_abstract}  
\input{sec/1_intro}

\input{sec/2_relatedWork}
\input{sec/3_method}

\input{sec/4_experiments}

{
    \small
    \bibliographystyle{ieeenat_fullname}
    \bibliography{main}
}

\input{sec/X_suppl}

\end{document}

%% file: sec/0_abstract.tex
\begin{abstract}
The development of Text-to-Video (T2V) generation has made motion transfer possible, enabling the control of video motion based on existing footage. However, current methods have two limitations: 1) struggle to handle multi-subjects videos, failing to transfer specific subject motion; 2) struggle to preserve the diversity and accuracy of motion as transferring to subjects with varying shapes.
To overcome these, we introduce \textbf{ConMo}, a zero-shot framework that disentangle and recompose the motions of subjects and camera movements. ConMo isolates individual subject and background motion cues from complex trajectories in source videos using only subject masks, and reassembles them for target video generation. This approach enables more accurate motion control across diverse subjects and improves performance in multi-subject scenarios. Additionally, we propose soft guidance in the recomposition stage which controls the retention of original motion to adjust shape constraints, aiding subject shape adaptation and semantic transformation. Unlike previous methods, ConMo unlocks a wide range of applications, including subject size and position editing, subject removal, semantic modifications, and camera motion simulation. Extensive experiments demonstrate that ConMo significantly outperforms state-of-the-art methods in motion fidelity and semantic consistency. The code is available at \url{https://github.com/Andyplus1/ConMo}.

\end{abstract}

%% file: sec/1_intro.tex
\section{Introduction}
Text-to-Video (T2V) generation \cite{ku2024anyv2v,yang2024direct,liang2025movideo} has advanced significantly with the evolution of video diffusion models \cite{gupta2025photorealistic,bar2024lumiere}. However, due to the inherent complexity of motion, current models struggle to control object dynamics and movement effectively. Thus, zero-shot motion transfer is proposed. Given a video description and a reference video, it aims to generate a video matching the description while preserving the reference video's motion patterns.

Previous methods \cite{kara2024rave,ling2024motionclone,tu2024motioneditor,wu2023tune,park2024spectral} often employ dense depth maps or sketches from reference videos to replicate specific motions. However, these motion cues are highly entangled to structural elements such as object shapes and scene layouts. To overcome this, recent approaches capture unstructured motion cues by aligning temporal attention maps \cite{ling2024motionclone,jeong2024vmc} or modeling per-frame differences \cite{yatim2024space} to guide the denoising process of target videos. Despite substantial progress, they still face two issues: (1) They produce suboptimal results in complex videos with multiple subjects motions. (2) When the generated subject differs from the original subject shape (especially with drastic difference), adapting the motion is challenging.
This is due to their use of a holistic motion representation from the reference video, which entangles compound motions from different subjects and the camera. 
In addition, the intensity of the original subject's motion in coupled motion guidance is uncontrollable in current approaches, leaving little flexibility for pretrained diffusion models to synthesize smooth transitions, particularly when the shape needs to change in accordance with the semantic content.

To tackle this issue, we introduce ConMo, a novel zero-shot framework to controllably transfer motions. Our key idea is to first \textbf{disentangle} compound motions in the reference video into individual subject and background motions, and then controllably \textbf{recompose} them during target video generation. Specifically, subject masks are first applied to calculate local space-time intermediate feature differences during inversion of the reference video. These differences serve as distinct motion cues specific to each subject. 
In addition, experiments proved that solely employing features derived using background mask can approximate camera movement trajectories \cite{hu2024motionmaster,cheong2024boosting}, as shown in Figure \ref{fig:teaser}(f), enabling more flexible and boarder applications.
Enlightened by this, we propose a soft guidance strategy that allows greater flexibility to alter subject category and shape. Specifically, by leveraging background motion to weight subject motion, we find that the resulting diluted subject motion offers greater flexibility for shape changes( as shown in Figure \ref{fig:teaser} (b), to maintain the same “right-to-left” motion, a “dog” would require an additional action like “run” compared to a “car”). Its effectiveness may stem from the introduced background motion that initially exits in the subject motion, which can reduce the shape-related constraint in the original object motion while maintaining overall motion harmony. 


Our ConMo can better handle complex motions patterns from multiple subjects by disentanglement and recomposition. 
By simply controlling the intensity of subject and background motion guidance through soft guidance, we achieve more adaptable subject alterations. Moreover, the recomposition strategy enables significant changes in the shape, position, and semantics of the subjects, as shown in Figure \ref{fig:teaser}(a-d). And we can also replace subject-specific motion cues with background one to remove subjects and simulate camera movement, as shown in Figure \ref{fig:teaser}(e-f). Our contributions are as follows:
\begin{itemize}
    \item We propose ConMo, a zero-shot framework for controllable motion transfer. Our method begins by disentangling compound motion into distinct subject and camera motions, which are then recomposed during target video generation.
    \item We propose a soft guidance strategy 
    to help recomposing more flexible motion dynamics. We first achieved fine-grained control over subject presence, size, position, and motion intensity.
    \item Extensive experiments demonstrate that ConMo achieves effective motion disentanglement and compositionality, outperforming previous methods in complex videos.
\end{itemize}

%% file: sec/2_relatedWork.tex
\section{Related Works}

\noindent \textbf{Video Motion Control.}
Extensive efforts have been made to customize generated video motions to align with user-provided text and other inputs \cite{yin2023dragnuwa, wang2024videocomposer, wang2024motionctrl, guo2025sparsectrl, ling2024motionclone, esser2023structure, chen2023control, shi2024motion, he2024cameractrl, haoran2023fine, zhang2023fine, ye2022unsupervised}. DragNUWA \cite{yin2023dragnuwa} presents a video generation technique that leverages text prompts, an initial image, and designated point trajectories. MotionCtrl \cite{wang2024motionctrl} facilitates precise control over camera poses and object motion, allowing for fine-grained motion manipulation. These approaches generally rely on training-based frameworks, demanding substantial training resources and often requiring additional modules to ensure that the generated videos conform to the specified conditions. Our method requires no additional training and utilizes easily obtainable masks as guidance. By disentangling video motion into subject and camera motions, it enables more precise motion control.

\noindent \textbf{Zero-Shot Text-Driven Motion Transfer.}
Text-driven motion transfer aim to generate a video that replicates the motion patterns from a reference video while enabling the generation of the target subject and scene through a text prompt. With the advancements of Video Diffusion Models\cite{zhang2024show, singer2022make, liu2024sora, molad2023dreamix, guo2024i2v} in video generation, current methods are capable of customizing these models to generate videos with tailored motions. Diffusion-Motion-Transfer (DMT) \cite{yatim2024space} extracts global feature differences to model motion cues, VMC \cite{jeong2024vmc} applies motion distillation in temporal attentions within a cascaded video diffusion. However, these methods  are designed to model overall motion, cannot handle motion transfer for videos with multiple subjects exhibiting individual motions.  MotionClone \cite{ling2024motionclone} employs sparse temporal attention weights as motion representations for motion guidance, Control-A-Video (CAV) \cite{chen2023control} uses control signals and a conditioning method based on the first frame. It extracts condition signals like sketches based on the input video and the structure of the generated video strictly follows the condition. They both remain restricted to motion transfer scenarios with limited shape transformation and struggle to handle complex movements. In contrast, our method is the first to extend motion transfer to the domain of multi-subject motion and introduces a strategy to decouple the motions of multiple subjects, while also supporting a broader range of shape transformations

%% file: sec/3_method.tex
\section{Method}
\label{sec:method}

\begin{figure*}[t!]
    \centering
    \includegraphics[width=1\textwidth]{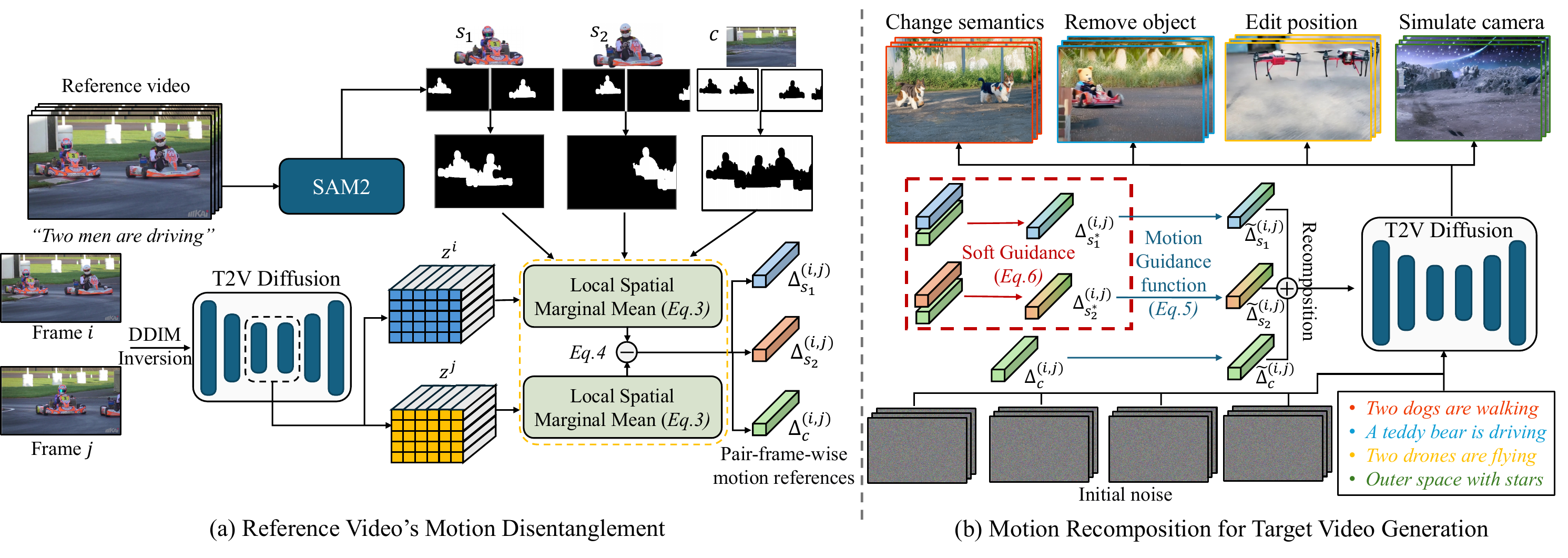}
    \caption{{\bf Overview of ConMo.} The method mainly consists of two stages:
    (a) Reference Video's Motion Disentanglement Stage: 
    We first acquire the masks for each subject in the reference video using SAM2\cite{sam2} and video latent features acquire during DDIM inversion\cite{song2020denoising}. Then, based on the mask, we identify the motion regions of each subject across different frames in the reference video. By calculating the difference of local spatial marginal means of latent features in these regions, we disentangle each subject’s motion. 
    (b) Motion Recomposition for Target Video Generation Stage: The extracted motion is integrated into the initial noise via the Motion Guidance function and Soft Guidance strategy. This allows generating target videos with consistent motion and adaptive shape handling. The method supports various video editing effects like semantic changes, object removal, position editing, and camera simulation.
    } 
    \label{fig:pipeline}
\end{figure*}

\subsection{Preliminary}

\noindent \textbf{Video Diffusion Sampling. }
Video Diffusion Models extend Latent Diffusion Models ~\cite{rombach2022high} by introducing temporal convolutions and cross-frame attention layers to model spatiotemporal dependencies, followed by video dataset fine-tuning for temporal consistency. Given an input sequence \( V = \{ x_0^1, x_0^2, \dots, x_0^n \} \), each frame \( x_0^i \) is first encoded into a latent representation \( z_0^i \). During the diffusion process, gaussian noise is progressively injected into \( z_0^i \) over timestep \( t \), yielding noised latents \( z_t^i \). The model then iteratively denoises \( z_t^i \) by predicting the noise component conditioned on temporal context, thereby reconstructing coherent video sequences through joint spatial-temporal learning.  

\subsection{Overall framework} 
As shown in Figure \ref{fig:pipeline}, ConMo operates in two stages: \textit{Reference Video’s Motion Disentanglement} and \textit{Motion Recomposition for Target Video Generation}. In the first stage, we disentangle subject-specific motions and camera motion from the reference video by extracting inverted latent features in pair-frame-wise dynamic regions associated with each subject via their corresponding masks. Then, it computes the difference of Local Spatial Marginal Means based on these features to represent independent motions for different subjects. In the second stage, the target videos are generated by recomposing the motions using the Motion Guidance function. It ensures target subjects' motion consistent with the reference subjects and adaptively handles different shape variations with Soft Guidance.

\subsection{Motion Disentanglement}
To achieve individual motion controls, we first disentangle the global motion cues of frame sequences into background motion and individual subject motions during reference video inversion process, as shown in Figure \ref{fig:pipeline} (a).
Unlike existing work \cite{yatim2024space} that only captures holistic motion cues by calculating mean differences of global features between frames, we identify locations of each subjects within video frames and compute local space-time features to model individual motion cues of separate subjects. 

Concretely, given the reference video $V$, we obtain the trajectory for a certain $k^{th}$ subject $s_k$ with mask $M_{s_k}$ using SAM2 \cite{sam2}. To acquire the local feature to model motion cue of $s_k$, for any two frames $i$ and $j$, such feature can be coarsely represented according to its changing region $M_{s_k}^i \cup M_{s_k}^j$ with Local Spatial Marginal Mean (LSMM):
\begin{equation}
    \phi({s_k}, i, j, t) = \frac{1}{\sum (M_{s_k}^i \cup M_{s_k}^j)}\sum f(z_t^i) \cdot (M_{s_k}^i \cup M_{s_k}^j)
    \label{eq:old_motion}
\end{equation}
where $f (\cdot)$ refers to extracting space-time features of $z_t^i$  from intermediate layers.

However, we find when multiple subjects move, especially when their trajectories overlap, the aforementioned process will result in the motion of the subject \( s_k \) to contain information from other subjects, making independent extraction impossible.
To overcome this issue, we propose to exclude the intersectional area between the current subject $s_k$ and other subjects $s_m$ in addition to $s_k$. 
Hence, the unique local motion region of $s_k$ across frames $i$ and $j$ is:

\begin{equation}
M_{s_k}^{i|j} = M_{s_k}^{i} \backslash M_{s_m}^j
\end{equation}

where $\backslash$ indicates the Set Difference function and the refined local feature for $s_k$ can thus be represented as:
\begin{equation}
\begin{split}
    &\phi({s_k}, i, j, t) = \\
    &\frac{1}{\sum (M_{s_k}^{i|j} \cup M_{s_k}^{j|i})}\sum f(z_t^i) \cdot ( M_{s_k}^{i|j} \cup M_{s_k}^{j|i} )
\end{split}
\label{eq:new_motion}
\end{equation}

The refined local feature $\phi({s_k}, i, j, t)$ in Eq.\ref{eq:new_motion} prevents interference from other subjects on the current motion cues, thereby providing more accurate space-time features for individual subjects compared to that in Eq.\ref{eq:old_motion}. Furthermore,
to obtain the isolated motion representation of \( s_k \) between frames \( i \) and \( j \), we compute the difference of local features of the same corresponding regions of the two frames:
\begin{equation}
    \Delta^{(i,j)}_{s_k} = \phi({s_k}, i, j, t) - \phi({s_k}, j, i,t )
\label{eq:difference}
\end{equation}

We also focus on the motions of background. Similar to previous procedures, the motion representations of background $c$ is denoted as $\phi(c, i, j,t)$. Empirically, we find that solely using background motions to generate video, the result mainly involves camera-changes, as shown in Figure \ref{fig:cameramotion}.

In summary, we disentangle the independent motion of subjects and camera from the reference video and acquire their motion representation in this stage, which serves as the foundation for subsequent motion recomposition for target video generation and other applications.

\subsection{Motion Recomposition with Soft Guidance}
To recomposite the individual motions to guide the generation of the target video. There are two requirements: 1. The subjects in the target video should follow the motion of the corresponding ones in the source video. 2. There should be more flexibility to handle more significant shape changes. Specifically, to ensure the target subjects preserving its corresponding reference motion, a guidance function is used to optimize latent features during the denoising process,  for individual subject $s_k$, the loss in each timestep $t$ is defined as:
\begin{equation}
   \mathcal{L}_{s_k}(f(z_t), f(\tilde{z}_t)) = \sum^n_i \sum^n_j \left\| \Delta^{(i,j)}_{s_k} - \tilde{\Delta}^{(i,j)}_{s_k} \right\|^2_2
\label{eq:energy_func}
\end{equation}
where $\tilde{z}_t$ and $z_t$ indicate latent variables from target denoising and reference video inversion process at the same timestep, respectively. (The frame indicator is omitted here for clarity.) And $\Delta$ and $\tilde{\Delta}$ are pair-frame-wise motion representations extracted from the reference video and target video, respectively, as calculated by Eq. \ref{eq:difference}.
For other subjects and background motion, we perform the same calculations to achieve recomposition.  This motion recomposition facilitates more accurate transition on complex motion videos with multiple subjects. It also makes it possible to add or remove a specified motion in reference videos, by simply adding or removing the corresponding guidance function.
However, due to strong semantic and geometric priors in extracted motion features \cite{yatim2024space, jeong2024vmc}, full-motion guidance without intensity control can limit flexibility in subject alterations, particularly for significant changes in semantics, shape, size or position. 
To address this, 
we propose a \textbf{soft guidance} scheme that blends subject motion with inherent camera trajectory control weights (as the subject moves within the camera view). This ``dilution" reduces the semantic / shape constraints of the original motion, enhancing the flexibility of our method.
Concretely, we compute the weighted arithmetic sum of the camera (background) motion and the subject motion as follows:
\begin{equation}
\Delta^{(i,j)}_{s^*_k} = \frac{\Delta^{(i,j)}_{s_k}+w_c * \Delta^{(i,j)}_{c}}{w_c+1}
\label{eq:soft_guide}
\end{equation}
where $w_c$ is a hyper-parameter to control the intensity of camera motion guidance. Larger $w_c$ values means weaker subject motion controls. Then we use new $\Delta^{(i,j)}_{s^*_1}$ to calculate energy function Eq.\ref{eq:energy_func} to guide motion generations. In this way, we achieve flexible motion controls to be more robust to shape and semantic changes.

\subsection{Applications}
With our proposed two strategies, ConMo enables a broader range of applications, as shown in Figure \ref{fig:teaser}. Below, we provide implementation details for each application:
\textbf{(1) Alter subject semantic and shape:} We can vary the degree of semantic and shape alterations for the subject by soft guidance (Eq.\ref{eq:soft_guide}) with different $w_c$. 
\textbf{(2) Control subject position and size:} We can adjust the position or size of the original motion in the generated video by resizing or shifting the mask in the corresponding area of $\tilde{\Delta}^{(i,j)}_{s_k}$. 
\textbf{(3) Remove motion:} we can replace $\Delta^{(i,j)}_{s_k}$ entirely with $\Delta^{(i,j)}_{c}$ to remove a specific motion.
\textbf{(4) Simulate camera viewpoint change:} By using only background motion cues $\Delta^{(i,j)}_{c}$ in the Eq. \ref{eq:energy_func}, we can simulate overall camera trajectory shifts.

%% file: sec/4_experiments.tex
\section{Experiments}

\begin{figure*}[t!]
    \centering
    \includegraphics[width=0.9\textwidth]{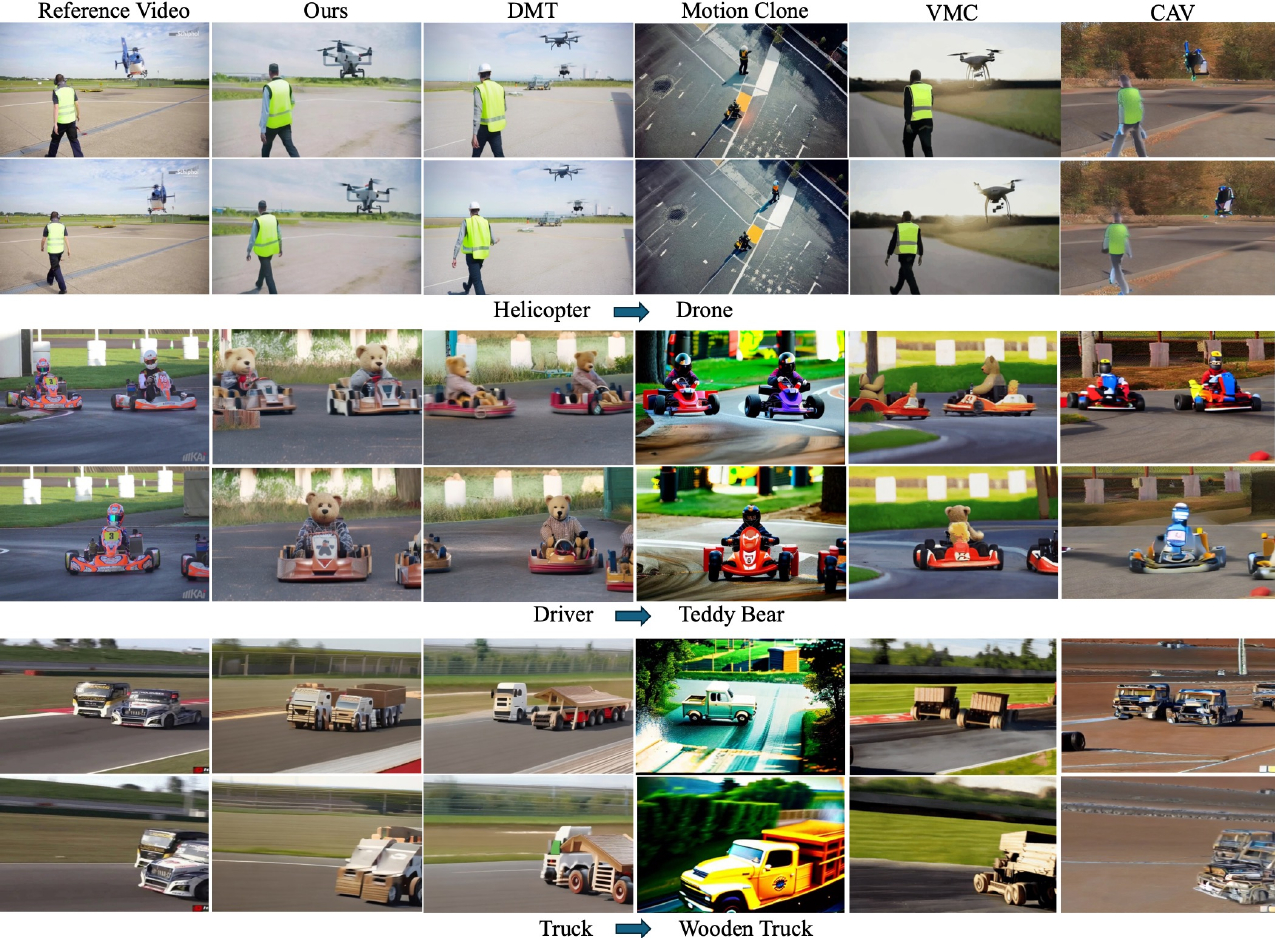}
    \caption{  \textbf{Qualitative Evaluation of multiple subjects motion transfer.} Our method achieves better results in term of text alignment and multi-subject motion fidelity.} 
    \label{fig:qualitative_multisub}
\end{figure*}

\begin{figure*}[t!]
    \centering
    \includegraphics[width=0.9\textwidth]{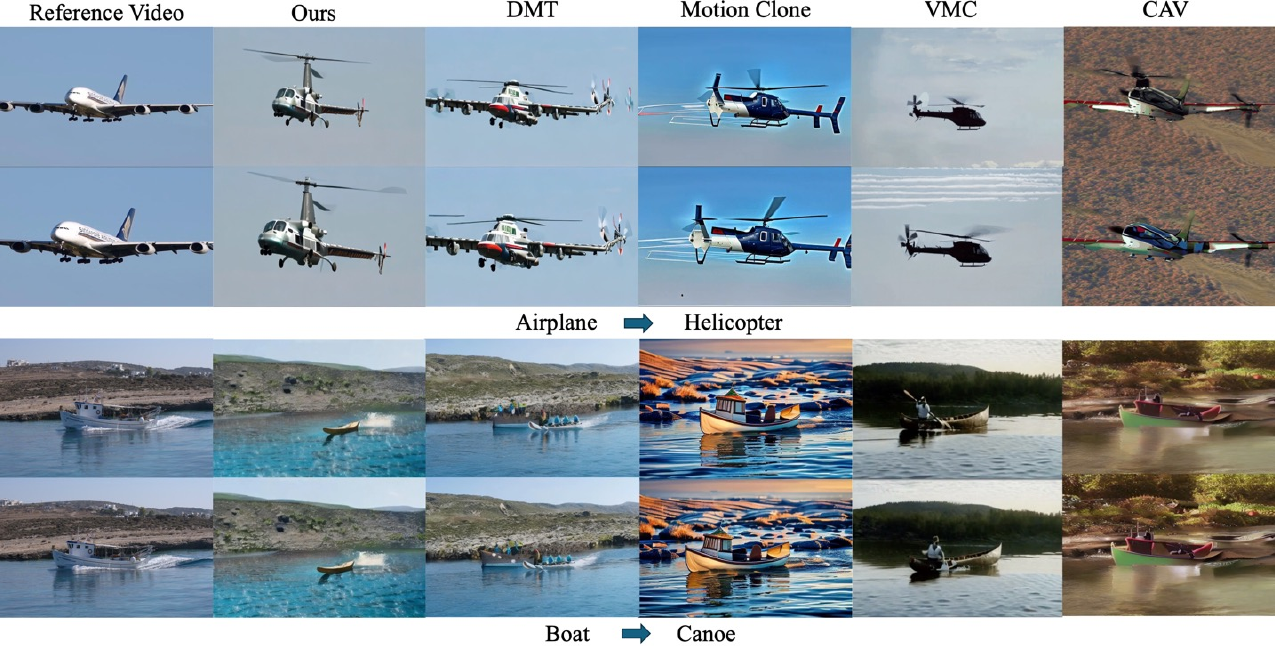}
    \caption{ \textbf{Qualitative evaluation of motion transfer with drastic semantic and shape alteration.} Our method outperforms other methods when subject shape changes are notable.} 
    \label{fig:shape}
\end{figure*}

\begin{figure}[t!]
    \centering
    \includegraphics[width=0.45\textwidth]{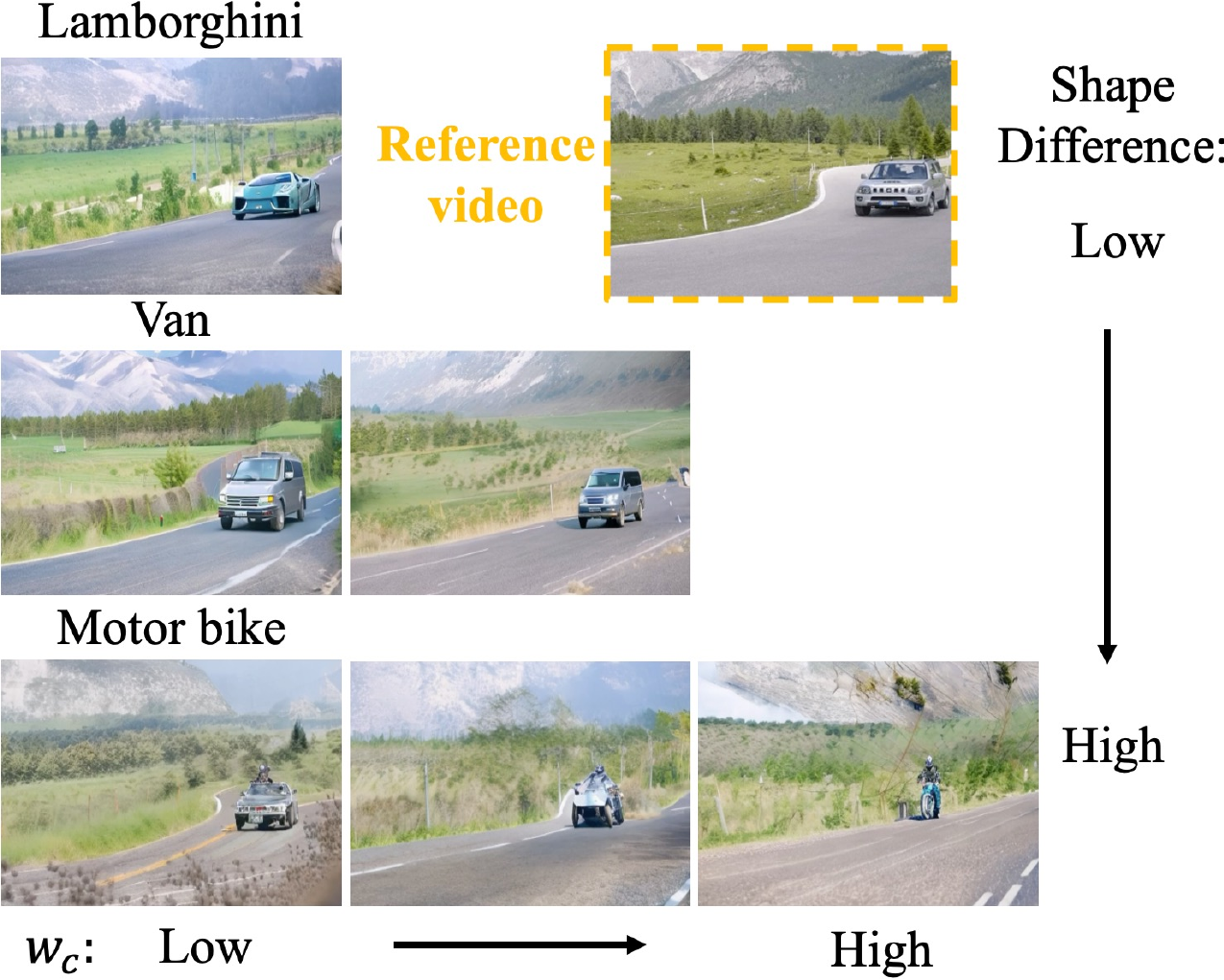}
    \caption{{\bf Controllable Motion Granularity.} Comparison of motion transfer across vehicle types with varying shape alterations. As background motion weight increases, original shape details diminish and alignment with prompts improves.
    } 
    \label{fig:shape_granularity}
\end{figure}

\begin{figure}[t!]
    \centering
    \includegraphics[width=0.45\textwidth]{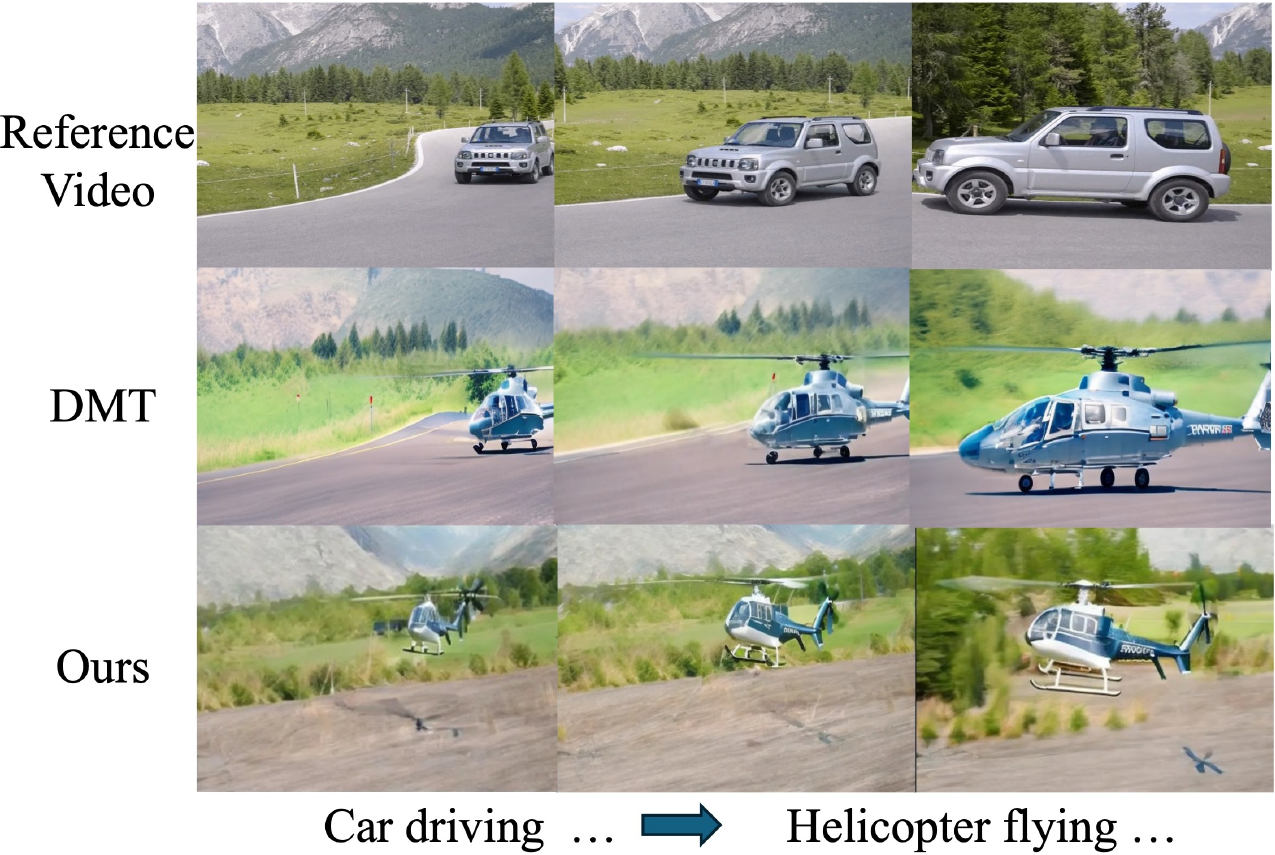}
    \caption{{\bf Reposition.}  By utilizing decoupled motion features, our method enhances the control of transferred motion positions, resulting in video outputs that more accurately align with the prompt's semantics (The flying helicopter should be in the sky).
    } 
    \label{fig:reposition}
\end{figure}

To fairly evaluate our method especially in complex video motions, we specifically collected a set of videos from DAVIS \cite{pont20172017}, TGVE \cite{wu2023cvpr} and the Web. Our dataset consists of 26 videos and 56 edited text-video pairs. We ensure scenes and object categories diversity. More implementation details are provided in the supplementary material.

\subsection{Qualitative Evaluation}
We provide visual comparisons of our method against four comparison approaches in scenarios involving multiple subjects and shape adaptation.

\textbf{Multiple subjects motion transfer.} As shown in Figure \ref{fig:qualitative_multisub}, 
DMT \cite{yatim2024space} can produce semantically accurate results, but when multiple motions are present, it fails to restore them individually (the generated drone has almost no motion). Moreover, it struggles to distinguish which subject is performing the action in cases with multiple separate, leading to mixed outputs (e.g. three teddy bears and three karts). MotionClone \cite{ling2024motionclone} lacks sufficient understanding of the text descriptions, resulting in outputs that do not align with the text in the three tested videos and appear disorganized. 
VMC \cite{jeong2024vmc} can coarsely preserve the motion of the original multi-subject video, such as overall motion trajectories, but there is still room for improvement in understanding fine-grained motion, such as the specific rotation changes of a drone, the orientation of a teddy bear during the driving process and the angles of the wheels during motion. CAV \cite{chen2023control} heavily retains the outline of the original video, as seen in the drone result, which still resembles a helicopter’s canny outline. Overall, our method generates videos that better meet the requirements of the text description while preserving the reference motions.

\textbf{Semantic and shape alteration.} As shown in Figure \ref{fig:shape}, we further compare our method with the aforementioned evaluation methods, focusing on cases where there are significant shape changes in subjects before and after editing. It is evident that DMT \cite{yatim2024space}, MotionClone \cite{ling2024motionclone} and CAV \cite{chen2023control}, while preserving motion, also retain much of the appearance information, resulting in an overall outline and size that remain very close to the original video. 
Although VMC \cite{jeong2024vmc} successfully achieves shape transformation, it still suffers from semantic inconsistencies. For example, the helicopter’s orientation is incorrect, and a person appear when generating a canoe, possibly due to the structure of the boat’s bow in the original video.
In contrast, our method overcomes this limitation, successfully achieving motion replacement with significant shape changes while preserving the original motion.

\subsection{Quantitative Evaluation}
Following DMT \cite{yatim2024space}, we evaluate our method using the following metrics: (1) Text Alignment (higher is better): We use CLIP \cite{radford2021learning} to assess the similarity between each frame and the target text, following earlier research (e.g. \cite{hu2024comd,yang2024eva}), and report the average score.  (2) Motion fidelity (higher is better): We adopt Motion-Fidelity-Score proposed by \cite{yatim2024space}, which assesses motion fidelity in videos by comparing the similarity of unaligned long trajectories. 

We additionally conducted a user study with 25 participants to evaluate the effectiveness of ConMo and all comparison methods. The study primarily assessed three aspects: the motion retention between the input video and the generated video, the motion quality of the generated video and the alignment between the target prompt and the generated video. The survey utilized a rating scale from 1 to 5. See the supplementary
material for more details.

Qualitative results are shown in the Table \ref{tab:re}. Our method achieves better results compared to baseline methods by maintaining high fidelity to both the target prompt and the original motion. VMC \cite{jeong2024vmc} maintains high text alignment score but has poor understanding and transfer capabilities for motion details such as orientation and pose. MotionClone \cite{ling2024motionclone} performs excellently in motion fidelity score but has a low text alignment score. This is because it often tends to present a structure similar to the original video, leading to a mistaken evaluation of good motion preservation. 

\subsection{Ablation Study}
\begin{figure}[t!]
    \centering
    \includegraphics[width=0.48\textwidth]{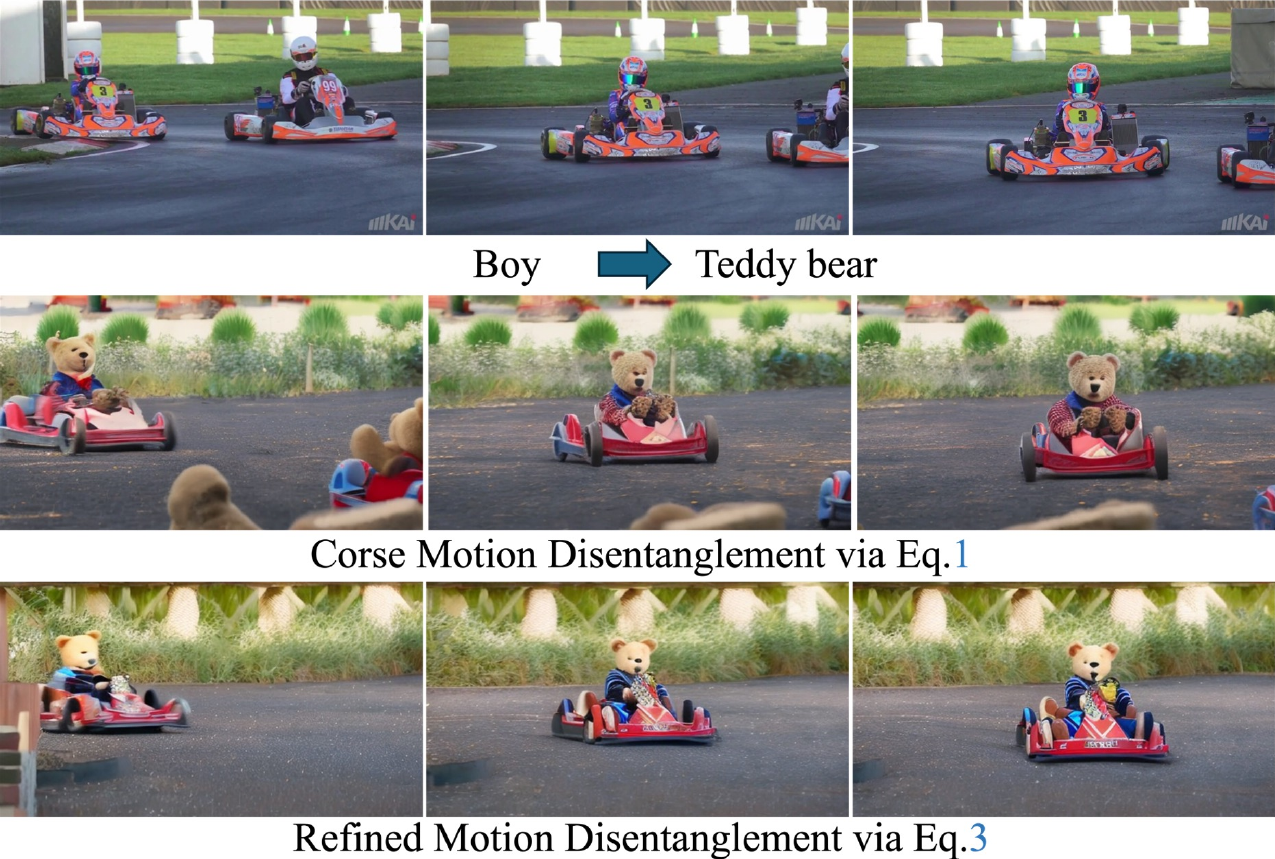}
    \caption{{\bf  Subject Motion Extraction.} The method effectively isolates the target subject's motion, ensuring independent transfer while excluding influences from other subjects. 
    } 
    \label{fig:subjectmotion}
\end{figure}

To validate the effectiveness of each module, we designed an experiment (as shown in Table \ref{tab:ab}). Compared to DMT's global guidance approach, our motion extraction method better preserves the original motion in the video. The application of Soft Guidance(SG) enhances the consistency between the generated video and the prompt, as it allows for greater shape transformation, making subjects better meet the prompt requirements. Additionally, we found it necessary to separate the movements of individual subjects with Eq.\ref{eq:new_motion}. This approach not only enhances model's performance but also demonstrates through visualization experiments that the extraction method can more independently isolate specific subjects' motion features. As shown in Figure \ref{fig:subjectmotion}, we use the single-subject motion extracted from multi-subject videos to guide video generation. Compared to the motion coarse extracted by Eq.\ref{eq:old_motion} (middle row), this method leaves residual motion from other subjects (such as the wheel on the right). The refined motion extraction by Eq.\ref{eq:new_motion} (bottom row) better preserves the independent motion, which lays the foundation for subsequent applications.

\begin{figure}[t!]
    \centering
    \includegraphics[width=0.45\textwidth]{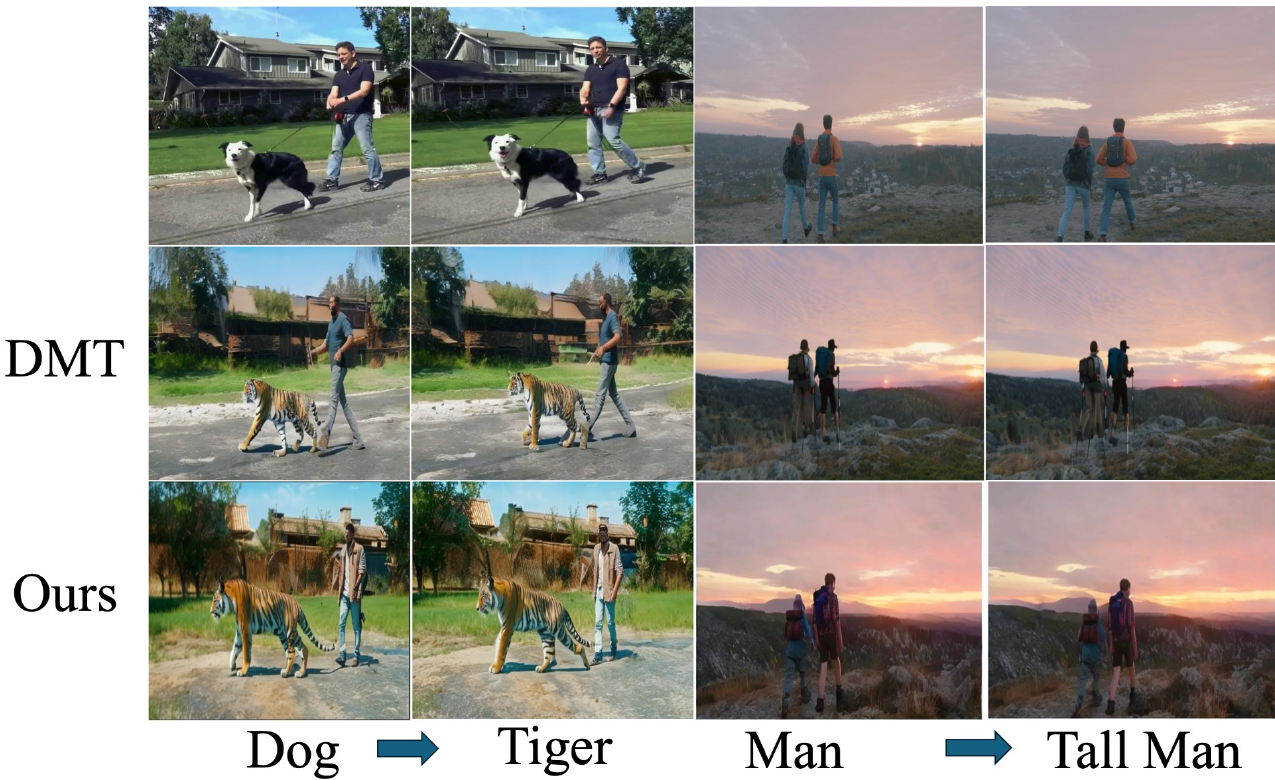}
    \caption{{\bf Resize.}
Our method precisely controls the size of generated subjects in videos, ensuring alignment with geometrically related text semantics and enhancing common sense alignment (e.g., a dog transforming into a larger tiger).
    } 
    \label{fig:resize}
\end{figure}

\begin{figure}[t!]
    \centering
    \includegraphics[width=0.5\textwidth]{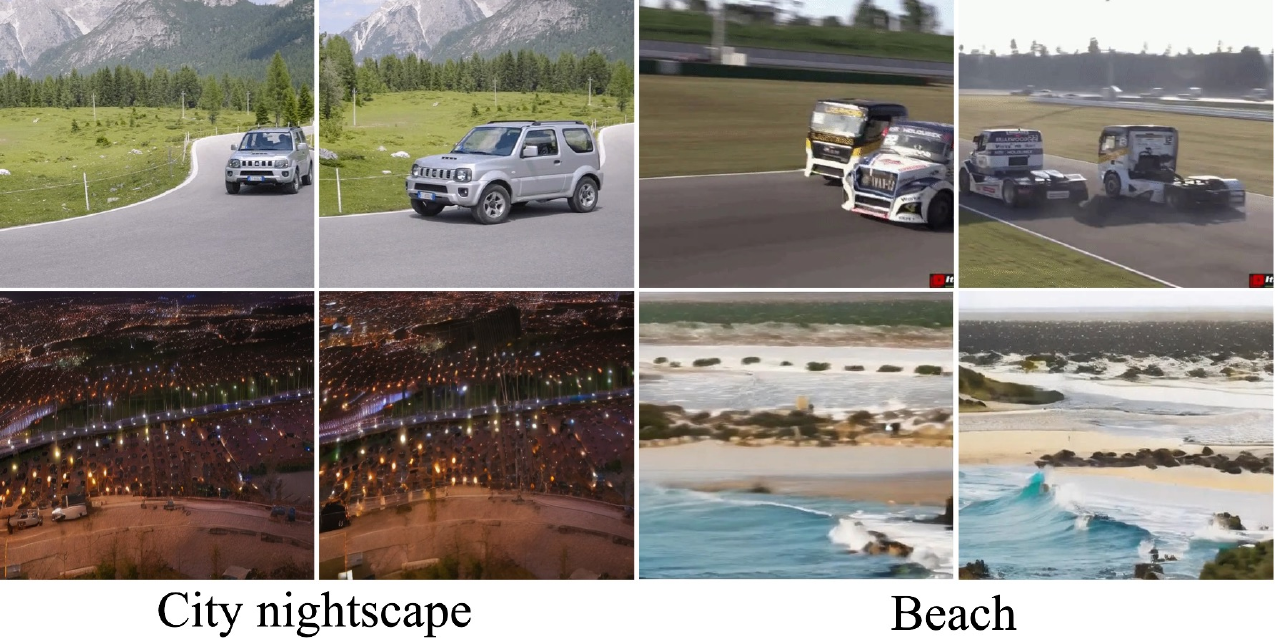}
    \caption{{\bf Camera Motion Extraction.}  The technique preserves original camera motion in the generated videos.
    } 
    \label{fig:cameramotion}
\end{figure}

\begin{table*}[]
\resizebox{\linewidth}{!}{
\begin{tabular}{lcc|ccc}
\toprule
Methods         & Text Alignment \textuparrow  & Motion Fidelity \textuparrow  & Motion Preservation \textuparrow  & Motion Quality \textuparrow  & Text Alignment \textuparrow \\ \hline
Control-A-Video  \cite{chen2023control}. & 30.13               & 0.7661           & 3.43                & 2,38            & 1.42             \\
VMC \cite{jeong2024vmc},            & \textbf{32.56}      & 0.7979           & 2.45                & 2.33            & 4.23             \\
MotionClone \cite{ling2024motionclone}     & 31.00               & 0.8876           & 4.20                & 3.40            & 3.01             \\
DMT \cite{yatim2024space}          & 31.46               & 0.8815           & 4.20                & 3.70            & 4.10             \\
\rowcolor{gray!40}
\textbf{ConMo}   & 31.96               & \textbf{0.8931}  & \textbf{4.40}       & \textbf{4.11}   & \textbf{4.30} \\ 
\bottomrule
\end{tabular}}
\caption{\textbf{Quantitative evaluations results with existing SOTA methods.} Evaluation results show ConMo considerably outperforms other methods in terms of motion fidelity, motion quality and text alignment, as demonstrated by automated metrics (\textit{left}) and user studies(\textit{right}).}
\label{tab:re}
\end{table*}

\begin{table}[]
\centering
\resizebox{0.9\linewidth}{!}{
\begin{tabular}{ccc}
\toprule
Methods  & Text Alignment \textuparrow  & Motion Fidelity \textuparrow  \\ \hline
DMT \cite{yatim2024space}       & 31.46          & 0.8675 \\
+Eq.\ref{eq:old_motion}     & 31.55          & 0.8813 \\
+SG       & 31.89          & 0.8795 \\\rowcolor{gray!40}
+Eq.\ref{eq:new_motion}      & \textbf{31.96}          & \textbf{0.8931} \\ \bottomrule
\end{tabular}}
\caption{\textbf{Ablation study}. Our proposed refined motion disentanglement and soft guidance strategy enhance model's performance.} 
\label{tab:ab}
\end{table}

\subsection{Applications}
\textbf{Alter subject semantic and shape.} 
As shown in Figure \ref{fig:shape_granularity}, we constructed prompts based on three different target subjects, each varying in morphology from the original subject. Increasing the weight of background motion enhances the alignment of the generated subjects with the prompt (e.g., Motor bike), while the retention of the original shape decreases. We attribute this to the injection of camera motion, which causes the motion of the original subject’s detailed structure to be gradually lost. (e.g., changes in the front of the vehicle). Consequently, the generated subjects are less constrained by this motion, allowing for more effective alignment with the prompts in the resulting videos. 

\textbf{Edit subject position.}
The position control method based on disentangled motion features not only enhances the flexibility of the motion transfer process but also improves the alignment of the generated video's motion with the prompt. As shown in Figure \ref{fig:reposition}, while the DMT method successfully transfers motion from a car to an airplane, the generated video shows the airplane taxiing rather than flying due to position constraints. In contrast, our position control method generates a video that accurately reflects the ``fly" semantic, and the generated subject seamlessly integrates with the video context (e.g., the shadow in the video corresponds to the generated helicopter).

\textbf{Edit subject size.} 
As shown in Figure \ref{fig:resize}, our method allows control over the size of specific subjects generated in videos, enabling better correspondence with geometrically related prompt semantics. More importantly, when transferring motion to a subject that differs significantly in size from the original, this technique can adjust the size of the generated subject based on the user-specified scaling ratios, ensuring the generated video aligns more closely with common sense (e.g., a dog becoming a tiger typically increases in size). This is particularly important in multi-subject videos, where other subjects serve as reference. 

\textbf{Remove motion and Simulate camera change.} 
Figures \ref{fig:subjectmotion} and \ref{fig:cameramotion} demonstrate ConMo’s ability to extract various types of motion from videos in a relatively independent manner. In Figure \ref{fig:subjectmotion}, the mask extraction method outlined in our Eq. \ref{eq:new_motion} allows us to avoid including information from other subjects in the extracted motion. This enables the removal of a specific subject’s motion from multi-subject videos, achieving an object removal effect. 
In Figure \ref{fig:cameramotion} , the videos generated from the background motion extracted in single or multi-subject videos effectively preserve the camera viewpoint transformations of the original video, while maintaining a low structural similarity to it showing that the background-extracted motion can be approximated as camera motion cues. 

More experimental results for the aforementioned applications are provided in the supplementary material to further demonstrate our method’s effectiveness.
\section{Conclusion}
In this paper, we propose ConMo, a novel approach to overcome the limitations of current text-to-video motion transfer methods through innovative motion disentanglement and recomposition strategies. By decomposing compound motions into distinct subject and background dynamics, we offer subject-level motion control capabilities, enhancing adaptability in scenarios with multiple subjects and complex motion patterns. Our soft guidance strategy enables flexible adaptation for target subject with different shape variations. With simple yet effective recomposition strategy to generate target video, we enable broader applications, including subject removal, editing subject geometry attributes such as size and position, and simulating camera changes. Extensive experiments further show that ConMo outperforms existing methods in maintaining motion consistency and flexibility. 

\noindent \textbf{Acknowledgements.}
This work was supported by the grants from the National Natural Science Foundation of China (62372014, 61925201, 62132001, 62432001) and Beijing Natural Science Foundation (L247006, 4252040).

%% file: sec/X_suppl.tex
\clearpage
\setcounter{page}{1}
\maketitlesupplementary

\renewcommand{\thesection}{\Alph{section}}
\setcounter{section}{0}

In the supplementary material, we provide additional information and
experimental results relating to ConMo. We begin by providing more details about the experimental setup and user study (Sec.\ref{sec:1}). Then, we provide more experimental results comparing our method with our baseline DMT \cite{yatim2024space}, focusing on the following three aspects: Multi-subject motion transfer, Fine-grained motion transfer and motion transfer with significant changes in shape (Sec.\ref{sec:2}). In Sec.\ref{sec:3}, We present additional results about applications focusing on repositioning and resizing. Finally we discuss the limitation of our method regarding the use of masks (Sec.\ref{sec:4})

\section{Implementation Details and User Study.} \label{sec:1}
\textbf{Training details:} To ensure a fair comparison with DMT \cite{yatim2024space}, we use the same parameter settings and feature selection. For the initial noise, we use the same initialization method as in DMT, which involves downsampling and upsampling operations, except for the resize and reposition processes, where we use randomly initialized noise.

\noindent\textbf{User study details:} For the user study on the right side of Table 1 in main manuscript, we investigated 25 participants to evaluate the effectiveness of ConMo and all the comparison methods on our dataset consists of 26 videos and 56 edited text-video pairs. 
The user study on the right side of Table 1  in main manuscript primarily assessed three aspects referencing VMC \cite{jeong2024vmc} and MotionClone \cite{ling2024motionclone}: the motion retention between the input video and the generated video, the motion quality of the generated video and the alignment between the target prompt and the generated video. The survey utilized a rating scale from 1 to 5. To evaluate motion preservation, the participants were asked: `` To what extent is the motion from the input video retained in the generated video? " To assess motion quality, participants were asked: `` Is the motion in the generated video sufficiently smooth? " To decide text alignment, participants were asked: ``Does the generated video semantically align with the target prompt? ” The result of Table 1 in main manuscript shows that our method outperforms the baselines in all three aspects.

\section{More Results Comparing with DMT} \label{sec:2}
In this section, we further illustrate our method through additional visualizations, primarily comparing it with our baseline DMT\cite{yatim2024space}.

\begin{figure*}[t!]
    \centering
    \includegraphics[width=1\textwidth]{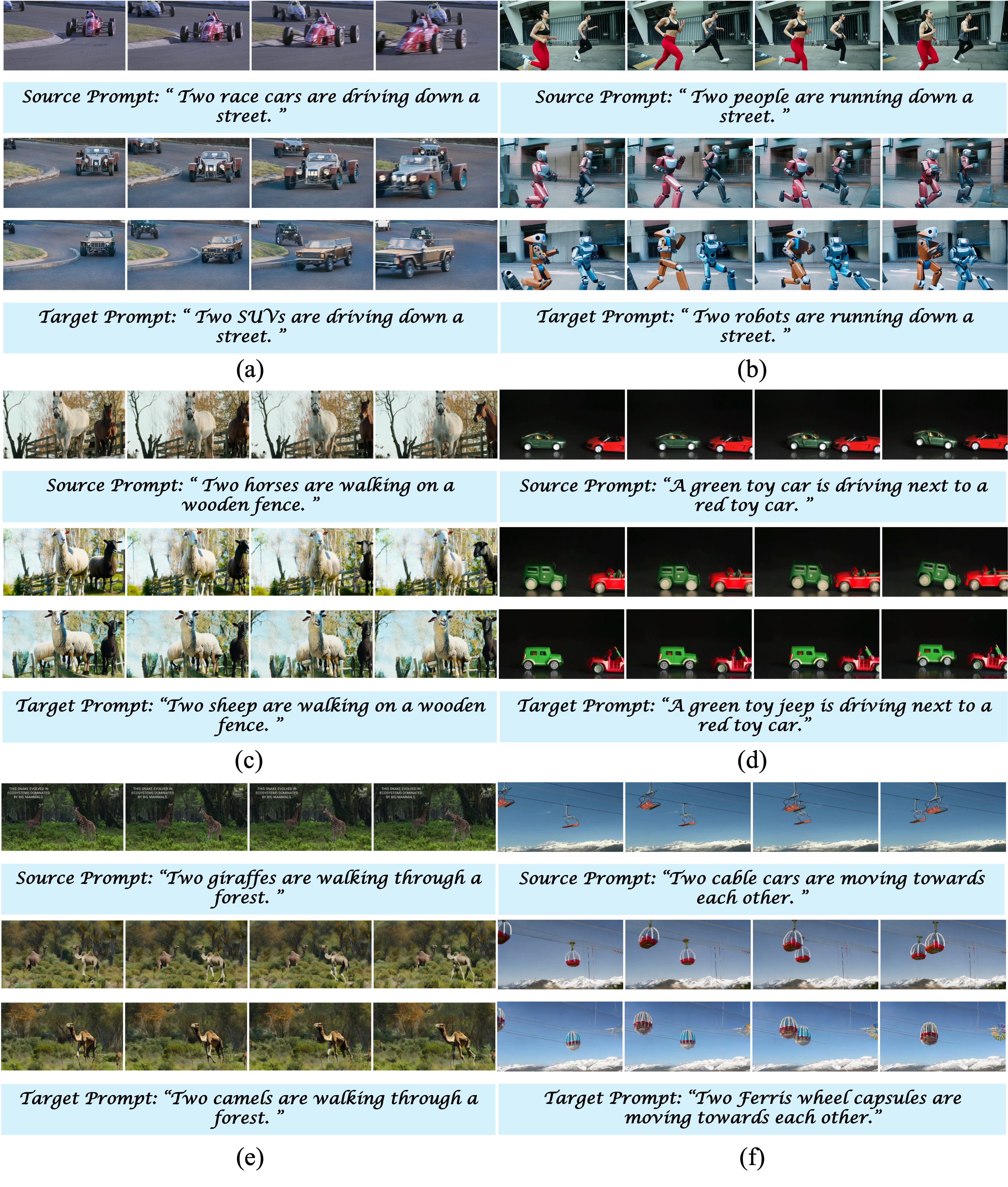}
    \caption{ {\bf Multi-subject motion transfer.} 
    We validate that our method achieves better motion retention for multi-subject videos. In each example, the results in the second row are from ConMo, and the results in the third row are from DMT \cite{yatim2024space}.}

    \label{fig:re_multi}
\end{figure*}

In Figure \ref{fig:re_multi}, We compare our method with the results generated by DMT\cite{yatim2024space} on multi-subject videos. In case (a), DMT\cite{yatim2024space} preserves holistic motion patterns but fails to distinguish individual subject trajectories  when two cars share identical motion in the source video, it erroneously generates additional vehicles along the common trajectory rather than establishing precise correspondence between the synthesized SUV and reference race car. This limitation becomes more evident in case (b) involving fine-grained limb movements, where DMT's motion extraction strategy \cite{yatim2024space} based on compressed global feature only retains dominant foreground actions (the woman's motion) with degraded articulation details, whereas our decoupling strategy successfully preserves nuanced limb dynamics across all subjects. When handling conflicting motions as shown in (c), DMT's \cite{yatim2024space} entangled motion representation collapses into static outputs when reference subjects exhibit opposing movements, while our approach accurately reconstructs the collision physics through separated motion modeling. Furthermore, in scenario (e) containing subjects with varying motion saliency, DMT\cite{yatim2024space} tends to suppress subtle movements of less active subjects, whereas our separated representation learning ensures simultaneous preservation of both prominent and latent motions through explicit motion decomposition. Beyond these cases, our method consistently outperforms DMT\cite{yatim2024space} across other examples in terms of video quality and robustness, with significantly fewer visual distortions and artifacts.

\begin{figure*}[t!]
    \centering
    \includegraphics[width=1\textwidth]{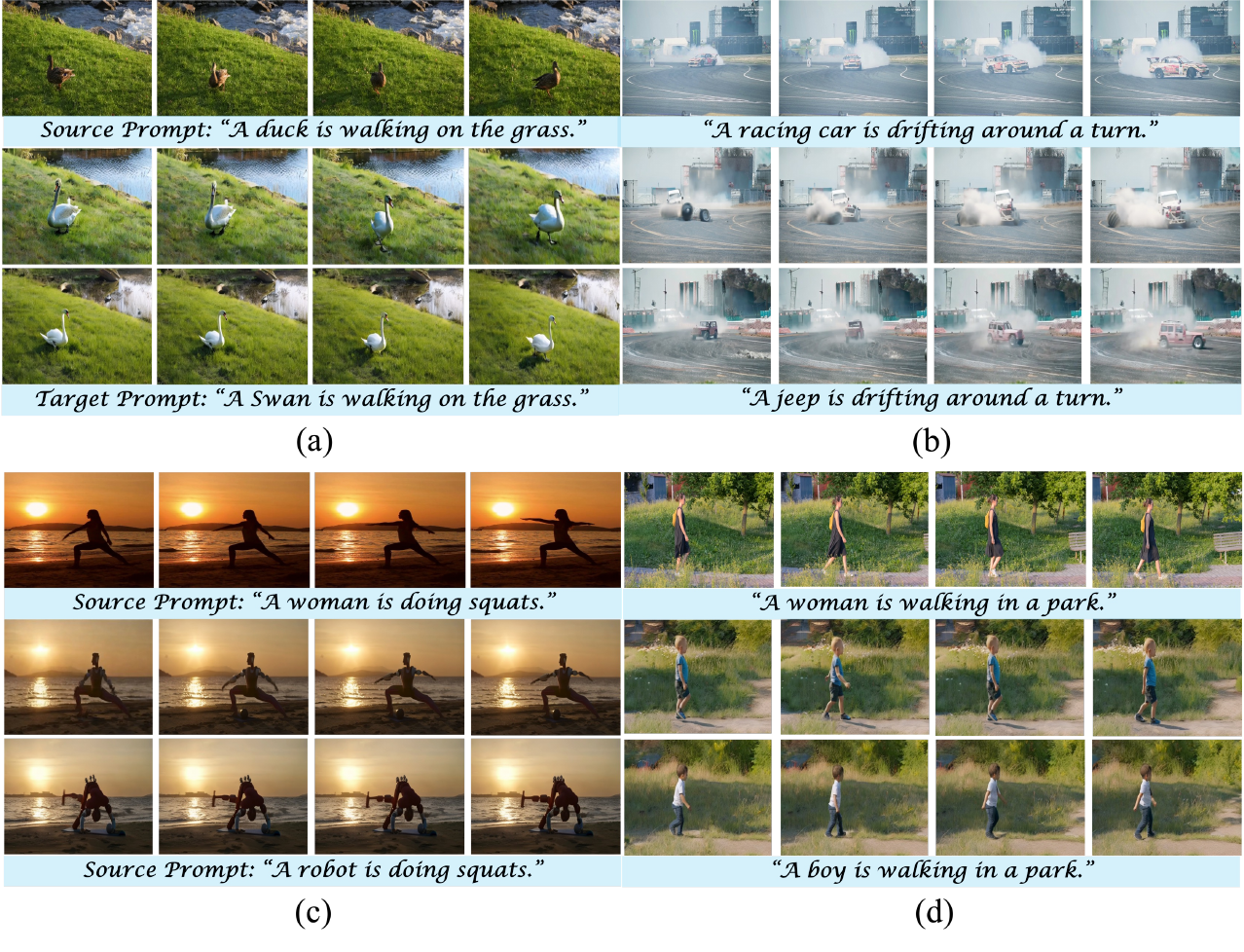}
    \caption{ {\bf Fine-grained motion transfer.} 
    We demonstrates that our method effectively maintains fine-grained motion. In each example, the results in the second row are from ConMo, and the results in the third row are from DMT\cite{yatim2024space}.} 
    \label{fig:results_finegrained}
\end{figure*}

In Figure \ref{fig:results_finegrained}, we compare our method's ability to preserve the original video's fine-grained motion against DMT\cite{yatim2024space}. 
In case (a), the duck's inconsistent motion direction and brief initial left-down motion cause DMT, which calculates motion globally, to overlook this process. In contrast, our method, which uses a fine-grained mask based approach, better retains the trajectory details. As a result, the generated video accurately preserves this part of the reference motion.
In case (b), the smoke's motion in the original video affects the global motion extracted by DMT\cite{yatim2024space}. This leads to the car's left-turn process being ``counteracted". The generated video shows the car moving in a straight line with many artifacts. In comparison, our method extracts the original drifting motion of the race car independently and transfers it well to the generated video. For cases (c) and (d), our method better preserves fine-grained human limb movements than DMT\cite{yatim2024space}, whose results appear unnatural.

\begin{figure*}[t!]
    \centering
    \includegraphics[width=1\textwidth]{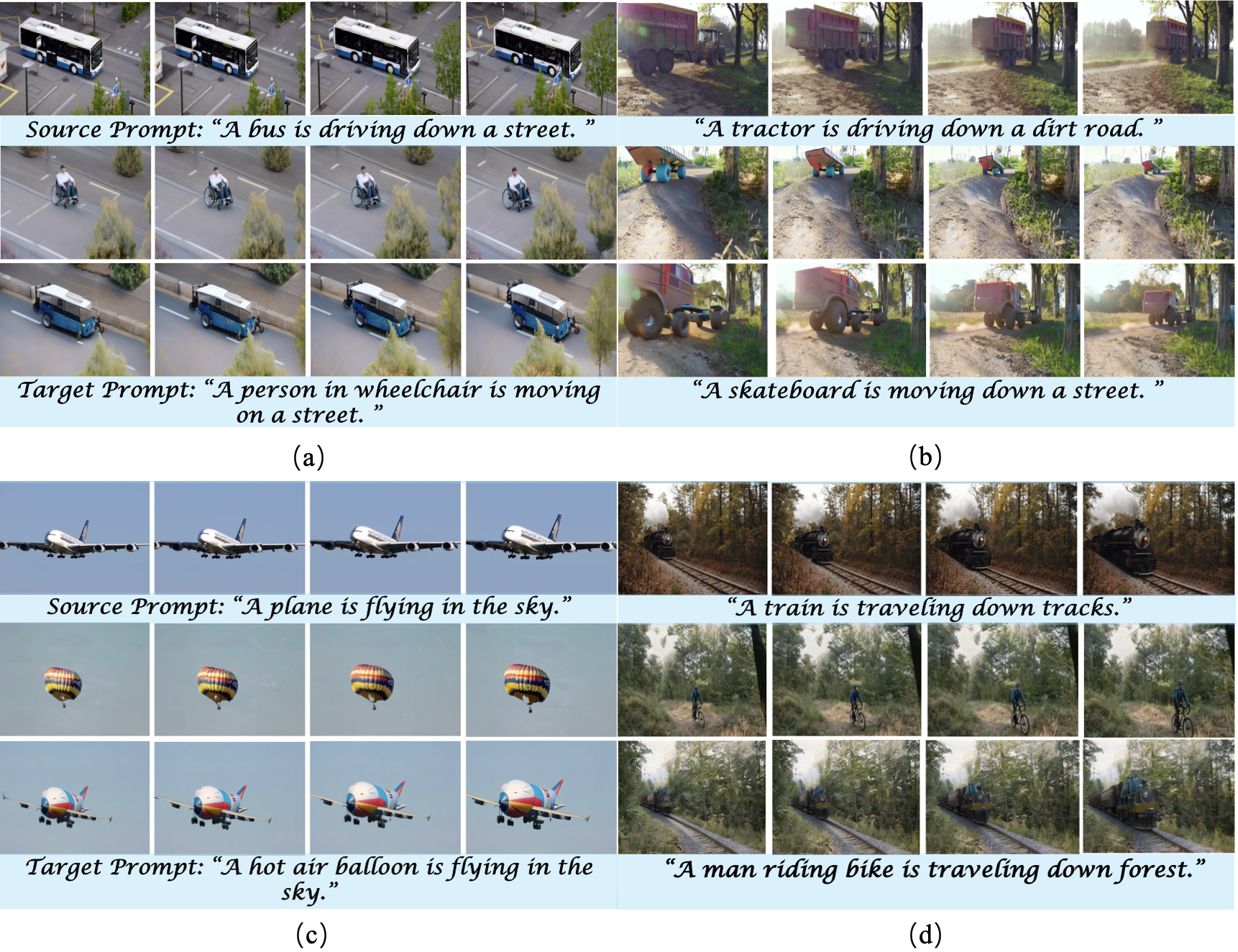}
    \caption{ {\bf Motion transfer with significant changes in shape.} 
    We demonstrates the motion transfer results of ConMo compared to DMT \cite{yatim2024space} when there is a significant difference in shape between the target subject and the reference subject. In each example, the results in the second row are from ConMo, and the results in the third row are from DMT \cite{yatim2024space}.
    } 
    \label{fig:results_bigshape}
\end{figure*}

In Figure \ref{fig:results_bigshape}, we further demonstrate that motion can be transferred to subjects with very drastic shape changes (such as from an airplane to a hydrogen balloon, from a train to a person riding a bicycle, etc.) through soft guidance with larger $w_c$. In contrast, DMT\cite{yatim2024space} is limited by the shape-related information in the original motion. As a result, it often only achieves texture replacement for the generated subjects, failing to realize complete shape changes.

\section{More Results about Applications} \label{sec:3}
\begin{figure*}[t!]
    \centering
    \includegraphics[width=0.5\textwidth]{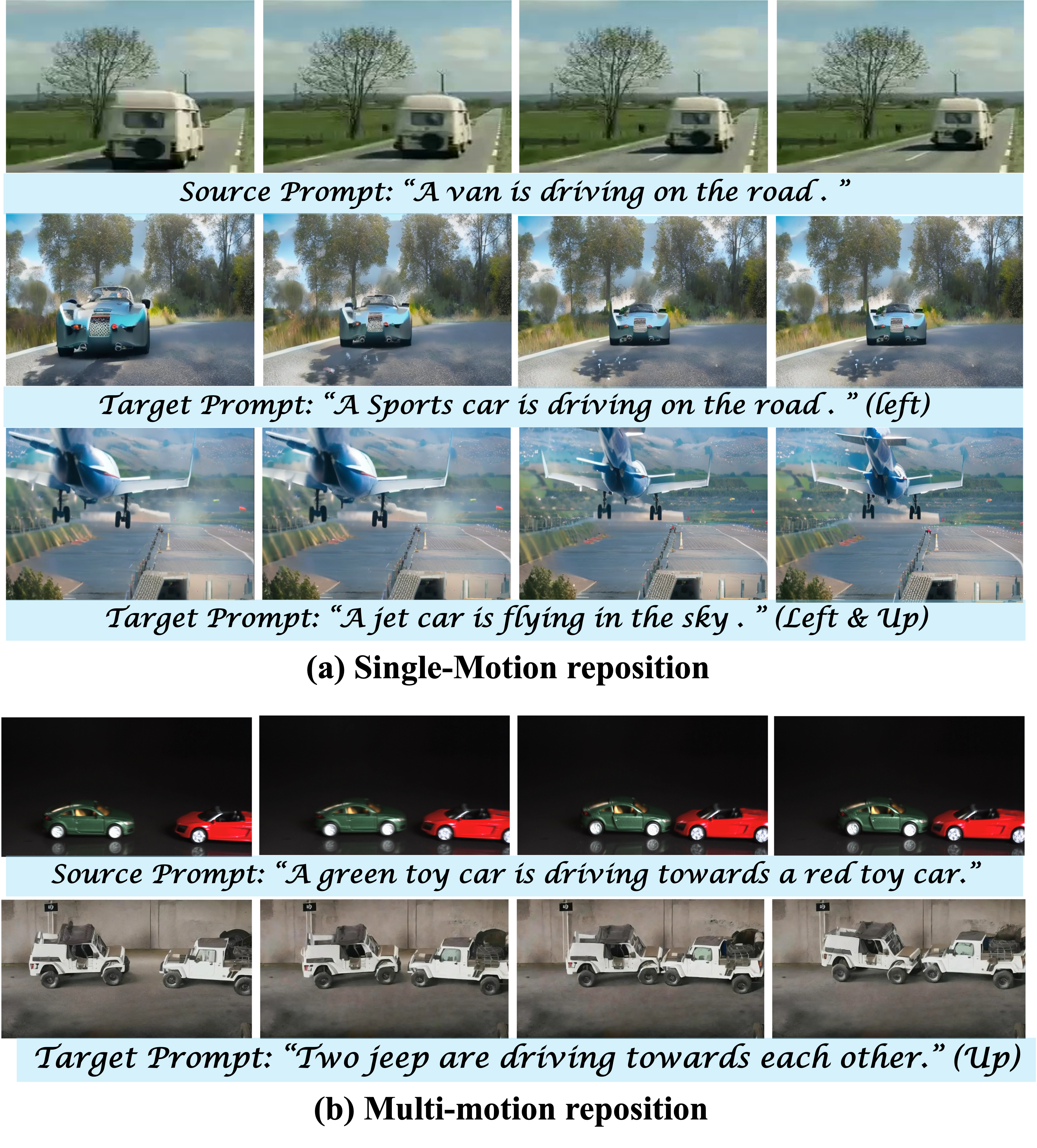}
    \caption{ {\bf Position Control.} 
    In (a), we have demonstrated our ability to reposition the main subject to a specified location (moving left and up), and as shown in (b), this operation can be applied to videos with multiple subjects. 
    } 
    \label{fig:results_position}
\end{figure*}

For the applications we proposed in the main text, we also present additional results here focusing on repositioning and resizing:

Regarding the repositioning task, as shown in Figure \ref{fig:results_position}, we have successfully achieved the horizontal and vertical movement of the original subject's motion, making the generated video more aligned with the target prompt's description. Moreover, we have demonstrated that the corresponding repositioning strategy can be transferred to videos with multiple subjects.

\begin{figure*}[t!]
    \centering
    \includegraphics[width=0.5\textwidth]{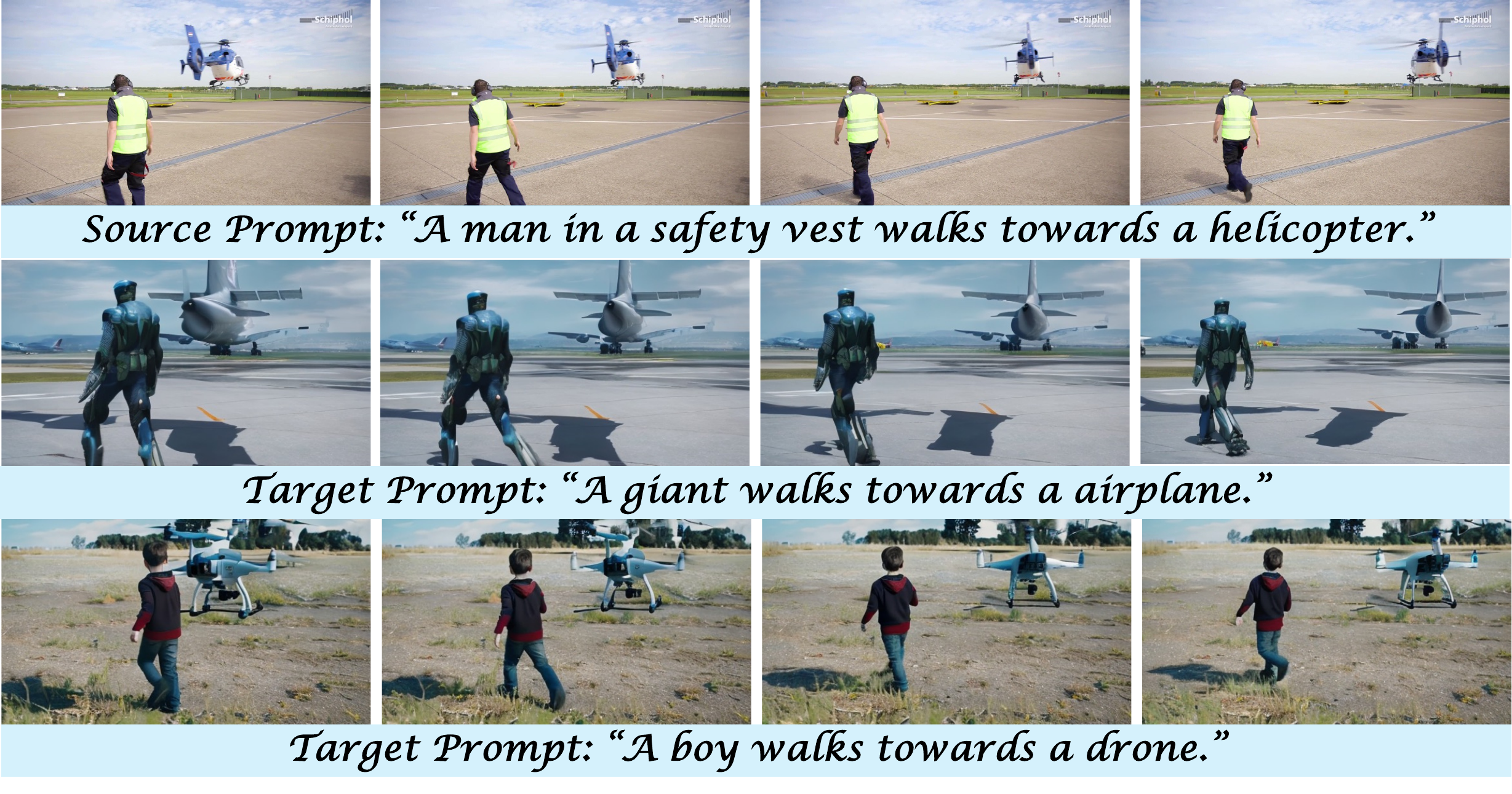}
    \caption{ {\bf Size Control.} 
    We have demonstrated our control capabilities over size, which allows the moving subjects in the video to present a more semantically appropriate effect (with 'boy' corresponding to a smaller size and 'giant' corresponding to a larger size).} 
    \label{fig:results_size}
\end{figure*}

For the resizing task, we further prove in Figure \ref{fig:results_size} that we can control the scaling of the target subject, from enlargement (man to giant) to reduction (man to boy), which is of significant importance for motion transfer that requires size control.

\section{Limitation} \label{sec:4}
\begin{figure*}[t!]
    \centering
    \includegraphics[width=0.5\textwidth]{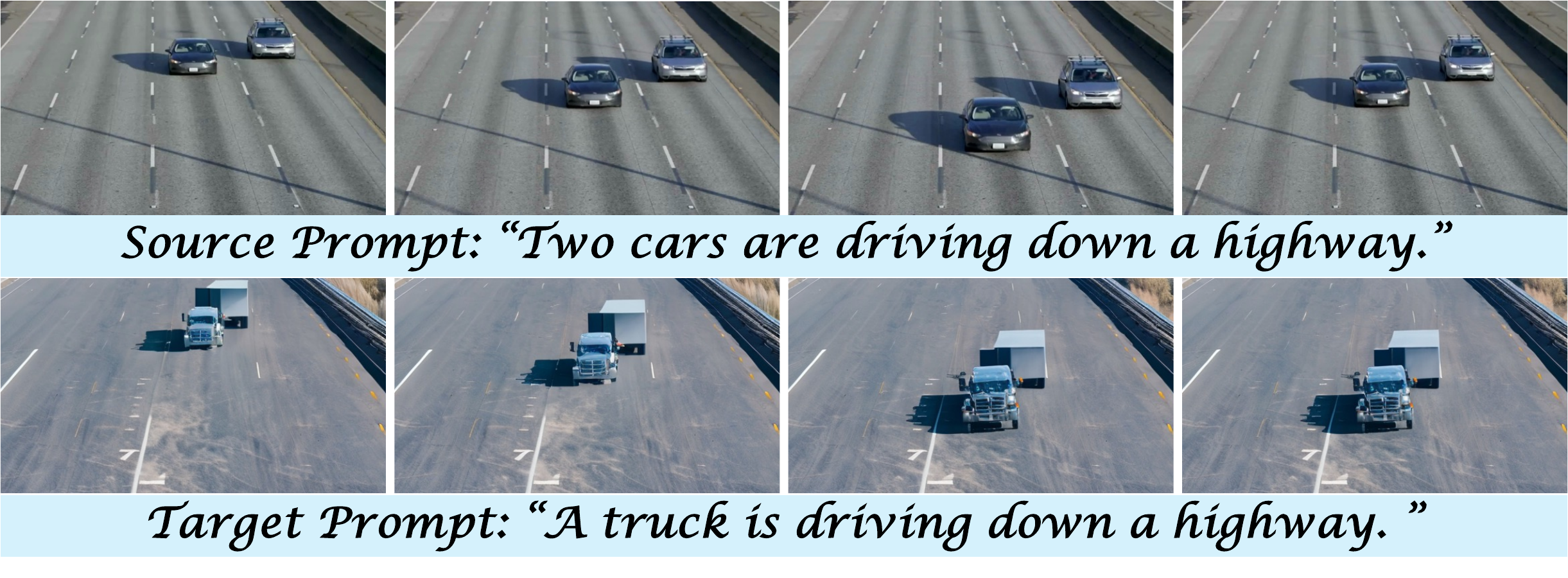}
    \caption{ {\bf Limitation.} In the process of removing the motion of the car on the left side of the original video, the segmentation model failed to account for the effects of the corresponding object, specifically the shadow in the video. As a result, the motion of the shadow can negatively impact the generated video.
    } 
    \label{fig:limitation}
\end{figure*}

Existing methods are limited by the mask segmentation process. If the mask input is incomplete or if the video contains effects caused by objects that cannot be annotated (e.g., large shadows), it may lead to the decoupled motion still containing information from other subjects, as shown in Figure \ref{fig:limitation}. Such contaminated motion can negatively impact the generated videos (causing artifacts, for example).

